\documentclass{vqa}

\title{Adventurer's Treasure Hunt:\\ A Transparent System for Visually Grounded Compositional Visual Question Answering based on Scene Graphs}

\addauthor{Daniel Reich}{dreich@uni-bremen.de}{1}
\addauthor{Felix Putze}{felix.putze@uni-bremen.de}{1}
\addauthor{Tanja Schultz}{tanja.schultz@uni-bremen.de}{1}

\addinstitution{
 Cognitive Systems Lab\newline
 University of Bremen\newline
 Bremen, Germany
}

\runninghead{Reich, Putze, Schultz}{Adventurer's Treasure Hunt}

\begin{document}

\maketitle

\begin{abstract}
% While raw performances on popular Visual Question Answering (VQA) datasets have seen impressive improvements in recent years, system transparency and visual grounding in the reasoning process remain aspects that many systems do not explicitly address
With the expressed goal of improving system transparency and visual grounding in the reasoning process in VQA, we present a modular system for the task of compositional VQA based on scene graphs. Our system is called “Adventurer’s Treasure Hunt” (or ATH), named after an analogy we draw between our model’s search procedure for an answer and an adventurer’s search for treasure. We developed ATH with three characteristic features in mind: 1) By design, ATH allows us to explicitly quantify the impact of each of the sub-components on overall VQA performance, as well as their performance on their individual sub-task. 2) By modeling the search task after a treasure hunt, ATH inherently produces an explicit, visually grounded inference path for the processed question. 3) ATH is the first GQA-trained VQA system that dynamically extracts answers by querying the visual knowledge base directly, instead of selecting one from a specially learned classifier’s output distribution over a pre-fixed answer vocabulary.

We report detailed results on all components and their contributions to overall VQA performance on the GQA dataset and show that ATH achieves the highest visual grounding score among all examined systems.

% We show that ATH achieves the highest visual grounding score among all examined systems. Furthermore, we report detailed results on all components and their contributions to overall VQA performance and show that ATH’s strong system transparency provides us with clear directions for future iterative improvements.
\end{abstract}

%-------------------------------------------------------------------------

\section{Introduction}
\label{sec:intro}
Visual Question Answering (VQA) \cite{vqa} deals with the multidisciplinary challenge of answering natural language questions about given images. As such, a VQA system requires proper handling of two types of inputs: 1) language, which encodes the query we seek to answer, and 2) images, which encode the search space for the query and act as knowledge base storing the answer. A third piece that completes this task is an inference engine that models the interaction between these two modalities and extracts the answer.

Efforts to integrate these three pieces have spawned a wide variety of modeling approaches in recent years \cite{lxmert, nmn, bottomup_paper, neural_state_machine}. While some systems focus on strong multi-modal representation learning using transformer-based models (\cite{lxmert, lu2019vilbert, 12in1}), others employ models that are trying to capture a more human-like inference process using multi-step reasoning, both in modular \cite{nmn, stacknmn, tbd, inferringprograms, pvr} and end-to-end approaches \cite{SAN, mac, mcb}. Developments to represent the image in a more structured way as a set of salient regions (i.e. objects detected by an object detector such as Faster R-CNN \cite{fasterrcnn}), as introduced in \cite{bottomup_paper, bottomup_tricks}, inherently improve explainability and visual grounding by enabling analysis of attention maps over a more symbolic representation of the image. This concept is refined to include relationships between detected objects by generating a scene graph that defines objects as nodes and relationships as edges \cite{neural_state_machine, relationaware, liang-etal-2021-graghvqa, lcgn}. In terms of visual grounding of an inference chain, structured representations of an image - like scene graphs - have the distinct advantage of supporting the creation of an image-grounded inference chain (or path) that is both easy to extract and intuitive to understand.

Structured representations like scene graphs have been shown to be particularly useful in compositional VQA (cVQA) \cite{neural_state_machine, lcgn, liang-etal-2021-graghvqa, explainablesg, neuralsymbolicvqa, hypergraph, pvr, regat}. In cVQA, a key characteristic of the question is its explicit reference to objects or regions in the image, as well as to their relationships with each other. We like to draw an analogy between the multi-step inference process in cVQA and an adventurer on a treasure hunt, who is closing in on a treasure by following certain clues, like descriptions of landmarks. When resolving a question in cVQA, each step (or clue) usually refers to certain image regions (or landmarks) and how they are connected with each other. With each step, we are taken closer to the question's target (“X” on the map), until we finally arrive at our destination and find the answer (or treasure). 
Approaching the VQA task in the manner described above intuitively requires strong object detection, attribute recognition, relationship detection, and a language parser that will parse the question into a multi-step query sequence directing us how to navigate the image. We believe, modeling the reasoning process in this explicit way can greatly enhance a systems capability for explaining its output as it sheds light on how it arrives at an answer.

\noindent\textbf{Our contributions:}
With the expressed goal to design a VQA system with strong system transparency and visual grounding of the inference process, we contribute the following:

1. A new modular VQA system, named ATH (“Adventurer’s Treasure Hunt”).

2. An in-depth evaluation of all ATH modules, including scene graph generation, which to our knowledge has not yet been quantified for GQA \cite{gqa_dataset} in the context of VQA systems.

3. We show that ATH has significantly stronger visual grounding compared to reference systems like MAC \cite{mac}, N2NMN \cite{n2nmn} and PVR\cite{pvr}.

4. To the best of our knowledge, ATH is the first GQA-based VQA system that does not select an answer from a specially learned softmax distribution over a pre-fixed answer vocabulary, but extracts it dynamically by querying the currently given scene graph, truly basing its answers on the contents of the knowledge base.

\section{Related Work}
\label{sec:related}

% related work
Various approaches exist that separate the reasoning process in a modular and explicit way by learning a program sequence that is subsequently executed on image features \cite{n2nmn, stacknmn, tbd, inferringprograms}. Our approach differs from these in particular in the choice of image representation: we operate on scene graphs and use symbolic features instead of (spatial) visual features.
Recent approaches increasingly employ scene graphs as image representation \cite{neural_state_machine, lcgn, liang-etal-2021-graghvqa, explainablesg, hypergraph, pvr, regat}. Our system shares many similarities with \cite{neural_state_machine} and \cite{neuralsymbolicvqa} in particular, but is a fully modular system (unlike \cite{neural_state_machine}) and operates on probabilistic scene graphs (unlike \cite{neuralsymbolicvqa}, which uses a discrete structural scene representation). 
All of these systems - with exception of \cite{neuralsymbolicvqa}, which queries a  discrete structural database for answers - have in common an answer production process that determines the output by means of a classifier over a pre-fixed answer vocabulary, and therefore differ from ATH in this aspect. 
The importance of generalization capabilities and visual grounding has been discussed and addressed in works such as \cite{vqacp, mutant, vqa_segmentation}, and quantified in the context of GQA in \cite{foundreason, pvr}, which we use as reference for some of our evaluations.

\section{Dataset}
\label{sec:dataset}

Throughout this work, we use the balanced set from GQA \cite{gqa_dataset}, which is a dataset created for compositional VQA on real-world images. GQA contains a cleaner subset of images and scene graphs from Visual Genome \cite{krishnavisualgenome}. Questions (and answers) were artificially generated. 

All model training we report on in this work is performed on the balanced train subset from GQA. We performed no cleaning of our own on the dataset. 

For object detection training, we note that the popular reference object detector from UpDn \cite{bottomup_paper} was trained using a heavily cleaned (unreleased) version of Visual Genome that is not identical to the GQA train dataset, making their performance results (unpublished in their paper, available on github \cite{bottomup_repo}) not fully compatible to ours. To the best of our knowledge, object detection performance on GQA has not yet been published in the context of VQA.

All evaluations in this work are performed on GQA's balanced val set, unless stated otherwise. Whenever we required a separate development set, we created one by randomly splitting off 10\% from GQA's balanced train set.

\section{System description}
\label{sec:system}

\begin{figure}
\includegraphics[width=\textwidth,keepaspectratio]{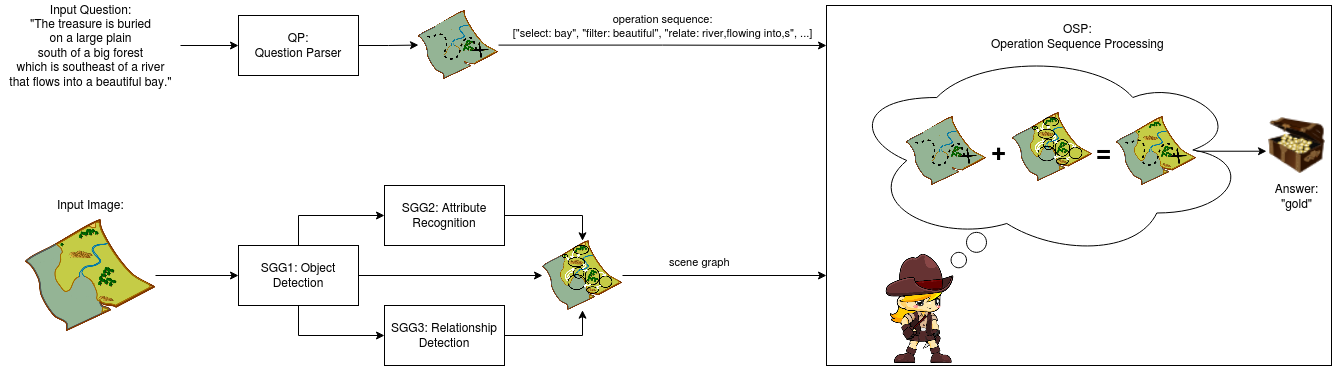}

\caption{Overview of ATH system components depicted as analogy of the Adventurer's Treasure Hunt. ATH consists of three main modules (QP, SGG, OSP), described in S\ref{sec:system}.)}
\label{fig:teaser}
\end{figure}

\subsection{Question Parser}
\label{sec:qp}

The Question Parser (QP) parses the question into a sequence of operations to be executed on the knowledge base (i.e. scene graph). These operation sequences (op-seqs) are part of the annotations of the GQA dataset, which we will use to train the QP. Operations follow a predefined structure given as “operation: argument“. Operations take on values such as “select”, “filter” and “relate”, while their arguments consist primarily of object, attribute and relationship names, as well as functional symbols like logical-or, underscore, etc.. We determined 136 unique operations. Due to highly variable nature of arguments, we pre-process them in a separate step before training the QP to generate them in a sequence. A post-processing step of the QP output is then required to revert the argument's (sub-)strings into the original format. Subsequent modules (namely OSP) process the op-seq in the original GQA format. Note that there are instances where this processing step might introduce issues to the final QP output, which can then cause problems downstream. We therefore isolate and report on the impact of this processing step in our evaluation (labeled "ATH*" in T\ref{table:refATHAblation} and T\ref{table:visualGrounding}). 

\noindent\textbf{Model description:}
The QP is based on a pointer-generator architecture with coverage presented in \cite{pointer_generator}. This network is, at its core, a seq2seq encoder-decoder model with attention mechanism with certain extensions that enable it to support "copying" a token from the input to the output sequence. We use this copy mechanism to avoid an unnecessarily large output layer when generating the op-seqs.

\noindent\textbf{Model specifications and Training:}
We use 50-dim, pre-trained Glove word embedding vectors \cite{glove} to encode words in the question. As recurrent layers in the encoder/decoder, we use a GRU \cite{gru} with 128 units. Question length is limited to 20 tokens which matches 99.5\% of our training data. The total question counts before removal are 849k (train), 94k (dev) and 132k (test). Output sequences in the test set are 7.4 tokens long on average. The softmax output layer of the model consists of 193 classes (20 classes for pointers to question words, 1 class to signal empty/no operation, 136 classes for operation terms, and the remainder for explicit modeling of the most frequently occurring (multi-)words in arguments, as well as non-word functional tokens like underscore, which cannot be copied from the input question). We use GQA's annotated pointers to determine whether or not an element in the target sequence should be a pointer to a certain question word in the input sequence. 
Output of our pointer-generator model is a pre-formatted op-seq. This sequence is a tokenized version of the original GQA op-seq, where tokens are defined as belonging either to the set of classes of the QP’s softmax output (in particular operation keywords like “select”, “filter”, etc.) or pointers to input question words (e.g. object names mentioned in the question that become arguments for operations as in e.g. “select table”, etc.).

We train our model with the Adam optimizer \cite{adam_optimizer} with 0.001 learning rate and batch size of 32 questions. Early stopping is applied with one epoch patience.

\noindent\textbf{Results: } The QP reaches 93.58\% accuracy on the test set (132k questions with avg output sequence of 7.4 tokens per question).

\subsection{Scene Graph Generation}
\label{sec:sgg}

The Scene Graph Generation (SGG) module is divided into three sub-modules, each responsible for a certain part in the modular construction of the scene graph. 
Output of the SGG module is the scene graph of the input image. Nodes represent objects and their attributes, edges represent relationships between objects. All instances (object ID, attribute ID, relationship ID) will be represented by each sub-module’s softmax distribution(s), which is an n-dim vector where n is the number of (accumulated) classes in the relevant sub-module.

We report performance for each sub-module on its own individual task, as well as its impact on the full VQA system in an ablation-type study (T\ref{table:refATHAblation}).
Find a description of each module in the following.

\subsubsection{SGG1: Object detection and visual feature extraction}
\label{sec:sgg1}
The heart of the SGG module is a Faster R-CNN \cite{fasterrcnn} model with ResNet101 \cite{resnet} backbone using a FPN \cite{fpn} for region proposals. We build the model using Facebook’s Detectron2 framework \cite{detectron2}. The ResNet101 backbone model was pre-trained on ImageNet \cite{imagenet_dataset}. 

\noindent\textbf{Training:}
We train the model for 1702 different object classes using 75k training images. Training lasts for 1m iterations with mini-batch of 4 images, using a multi-step learning rate starting with 0.005 and reducing by factor of 10 at 700k and 900k iterations. No other parameters were changed in the official Detectron2 training recipe for this model architecture. Training took about 7 days on a RTX 2080 Ti.

\noindent\textbf{Output:}
We extract the softmax output distribution (a 1702-dim vector) for each post-NMS detected object. Objects are selected as follows: per-class NMS is applied at 0.7 IoU for objects that have at least a max softmax object class probability of 0.05. We cap the output at 100 objects per image. We use the full softmax class distribution of each object that survived as features in the scene graph.

\noindent\textbf{Extraction of visual features for SGG2 and SGG3:}
As input features to ATH's modules for attribute recognition (SGG2) and relationship detection (SGG3), we extract 1024-dim visual features from the object classification head of the SGG1 model, just before the final fully connected softmax-activated output layer. This is done for each surviving object (i.e. after per-class NMS and top100 capping).

\noindent\textbf{Results:}
SGG1's mAP scores are as follows: 5.54mAP@[.5:.95]IoU; 9.45 mAP@0.5 IoU; 5.76mAP@0.75 IoU. For an approximate comparison: UpDn's \cite{bottomup_paper} object detection model achieved 10.2 mAP@0.5 on a heavily cleaned subset of Visual Genome with 1600 objects (unpublished, listed on github \cite{bottomup_repo}).

\subsubsection{SGG2: Attribute recognition}
\label{sec:sgg2}
We identified a total of 617 individual attribute names and 39 overarching attribute categories in GQA, including a category containing un-assigned attributes. Classifier sizes range from a maximum of 427 classes ("other" category) to a minimum of 2 classes (e.g. "height"), with most classifiers covering 3 or less classes. Categories and category membership were determined based on their declaration in GQA.
We train a separate softmax regression model for each of these 39 attribute categories.

\noindent\textbf{Training:}
Input to each model is a detected object’s 1024-dim visual feature vector extracted from SGG1. We assign category/attribute labels to a detected object when exceeding an IoU of 0.75 with a ref object. A sample is used to train every category model it has labels for.

\noindent\textbf{Output:}
Attributes for detected objects in the scene graph are created as follows: we feed a detected object’s 1024-dim visual feature vector into each of the 39 models and extract the softmax activation output distribution over each category’s classes. The resulting 39 distributions then represent the complete 617-dim attribute information of an object in the image’s scene graph.

\noindent\textbf{Results:} We report averaged results for attribute recognition over all 39 category models. Avg accuracy is 72.94\% (std: 14.22), weighted avg accuracy is 57.69\% (std: 14.49). Avg chance level (avg of largest class in each model) is 59.22\% (std: 20.16), weighted avg chance level is 30.61\% (std: 20.51).

\subsubsection{SGG3: Visual relationship detection}
\label{sec:sgg3}
We identified 310 relationship names in the dataset (e.g. “wearing”), including 13 spatial relationships (e.g. “behind”). For each category, we train an LSTM with 512 units, followed by a dropout layer (drop rate=0.3) and a softmax output layer.

\noindent\textbf{Training:}
Input to the LSTM are two object's 1028-dim vectors (1024-dim visual features, 4-dim bounding box coordinates, taken from SGG1) as a sequence in the order of [subject, object] according to their directed relationship. Samples are determined similarly to SGG2.

\noindent\textbf{Output:}
We classify relationships between objects that are sufficiently close to each other: We increase all object's bounding boxes by 15\% on each side and then select only those object pairs for classification that have any kind of overlap. 
Similar to SGG1\&2, we use a model's softmax output to act as relationship distribution between two objects and retain the subject-object order. As such, this probability distribution over all relationships serves as a weighted, directed edge in the scene graph.

\noindent\textbf{Results:} Spatial relationship detection (13 classes) achieves 95.77\% accuracy (chance: 46.92\%), non-spatial relationship detection (297 classes) achieves 69.49\% (chance: 32.64\%). There are far more spatial relationships annotated in GQA, which is reflected in our test set (197k spatial vs 13k non-spatial samples).

\subsection{Operation Sequence Processor}
\label{sec:osp}

The Operation Sequence Processor (OSP) takes as input the operation sequence (op-seq) from the QP and the scene graph from SGG and produces the answer to the question after traversing the scene graph. 
OSP addresses the following question: Given an op-seq and a scene graph, what is the most likely path in the scene graph that matches the path described by the op-seq? We draw inspiration from the decoding process in an automatic speech recognition system and use the Viterbi algorithm to handle this task. Class probabilities of objects, attributes, relationships given in the scene graph are used to determine the probability of nodes/objects matching the descriptions in the op-seq. Probabilities for nodes and edges are re-determined at each step - depending on the given operation in the op-seq - and are then used in the calculation of the Viterbi path.

Answers depend on the final node(s)/object(s) that we arrived at by following the Viterbi path(s). The extraction of the answer for "query"-type questions, is then a simple query on the scene graph for the maximum class in the object’s class distribution (e.g. object or attribute name as defined by the scene graph). 

We note that some questions contain two separate paths (e.g. logical-and type questions asking about existence of two separate objects). In these cases we process two Viterbi paths separately and apply the final operation on both path’s ending node to answer the question. 
“Verify”-type binary questions, that query object existence in the image, are answered by comparing the geometric average of the final best Viterbi path probability to a threshold value that was determined on a dev set (based on F1-score for thresholds on binary questions). 

We return the final Viterbi path in order to provide the user with a highly transparent view at the inner workings of the OSP and how ATH arrived at the answer.

One of the highlights of ATH lies in the way the answer is directly extracted from the knowledge base, supporting expansion of answer options without changing or retraining the inference system itself.

\begin{table}
\begin{center}
\begin{tabular}{|l|c|c|c|c|}
\hline
System & Binary & Open & Accuracy & Grounding \\
\hline\hline
N2NMN* \cite{n2nmn} & 72.59 & 40.30 & 58.94 & 55.44 \\
% UpDn \cite{bottomup_paper} & 61.98 & 38.80 & 50.02 & 70.10 \\
UpDn \cite{bottomup_paper} & 64.88 & 39.98 & 52.03 & 69.95 \\
% MAC & 77.90 & 48.37 (81.67) & 86.75 & 62.66 (79.85) \\
MAC \cite{mac} & \textbf{77.90} & \textbf{48.37} & \textbf{62.66} & 86.75\\
PVR* \cite{pvr} & 74.58 & 42.10 & 57.33 & 97.44\\
ATH & 58.55 & 37.23 & 47.55 & \textbf{125.63} \\
% ATH & 58.55 & 37.23 (63.59) & 125.63 & 47.55 (61.14) \\
\hline
\end{tabular}
\end{center}
\caption{Accuracy and grounding results, sorted by grounding. N2NMN* and PVR* accuracy results (reported in \cite{pvr}) are based on the GQA challenge test set (not used by us in this work). All system's grounding results are based on the GQA balanced val set. UpDn, MAC and ATH were trained by us and evaluated on GQA's balanced val set.}
\label{table:rawAcc}
\end{table}

\section{Experiments}
\label{sec:experiments}

In this section, we evaluate all parts of our system in detail. We note that our focus in this work is not on raw accuracy performance but rather on 
a) high system transparency for straightforward error blaming and analysis (see e.g. T\ref{table:refATHAblation}), b) strong visual grounding in the inference process (see e.g. T\ref{table:visualGrounding}), and c) an answer production process that truly bases its answers on the contents of the knowledge base (see also \ref{sec:osp}).

In order to place the system with other VQA systems, we evaluate a number of systems in T\ref{table:rawAcc}. We re-implemented UpDn \cite{bottomup_paper} and trained both UpDn and MAC \cite{mac} with ATH-produced 1024-dim region-based visual features. We trained MAC according to instructions at \cite{mac_github} (4-step inference for GQA, using region-based features instead of spatial features).

As shown in T\ref{table:rawAcc}, ATH's accuracy performance falls behind the reference systems, but surpasses all of them in visual grounding.

\subsection{ATH Module Evaluation}

\noindent\textbf{Operation Sequence Processing (OSP):}
We start by evaluating ATH using GQA annotations (=Oracle scene graph and op-seq). The upper-bound performance of ATH when running in full Oracle mode is 92.31\% accuracy (T\ref{table:refATHAblation},A-Oracle). Problems occur e.g. due to issues in the annotations or when processing exceptions that are not explicitly handled in OSP's mapping procedures (mapping operations to graph traversal operations) due to diminished return on investment.

\noindent\textbf{Question Parser (QP):}
Using an Oracle scene graph and a QP-\textit{processed} (not QP-\textit{predicted}) op-seq shows a drop in accuracy to 84.71\% (T\ref{table:refATHAblation},A8), indicating problems in QP's pre-/postprocessing of op-seqs. Closer analysis reveal in particular  conflicts w.r.t. word pointers created in QP that cannot be correctly resolved in post-processing. 
Using a fully ATH-created op-seq with Oracle scene graph shows another accuracy drop to 80.56\% (T\ref{table:refATHAblation},A7), which can be attributed to QP's classification errors.

\noindent\textbf{Object detection (SGG1):}
Object detection is by far the most influential component in terms of accuracy impact. If an undetected object is queried in the question, ATH's ability to arrive at the correct answer is heavily impaired. 

ATH's SGG1 detects 91.49\% of all objects in GQA's annotated inference chain, and 92.34\% of objects needed to answer a question (detection determined for IoU >0.5). This means that almost 8\% of questions cannot be reasonably answered correctly on account of critical objects missing in the scene graph. All numbers are listed in (T\ref{table:visualGrounding},Det.Obj.).

In addition to non-detections, having correct object class recognition for a large number of objects like in GQA (1702 object classes) is clearly challenging (see S\ref{sec:sgg1} for results). To evaluate the impact on ATH's overall accuracy, we create the scene graph by using SGG1's detected objects and retain all annotated attributes and relationships for detected objects matching the annotations ("matching": highest IoU >0.5). As expected, this heavily impacts ATH's overall accuracy which falls from 92.31\% to 69.23\% (-25\% rel.) for Oracle op-seq (T\ref{table:refATHAblation},A6), and 80.56\% to 63.13\% (-22\% rel.) for ATH op-seq (T\ref{table:refATHAblation},A5).

\noindent\textbf{Attribute recognition (SGG2):}
Similar to object detection, attribute recognition is important for identifying objects referenced in the op-seq. We replace the GQA-annotated attributes in the partial Oracle scene graph from SGG1 with outputs from the SGG2 module. 
Although the impact on ATH's accuracy here is not as strong as for SGG1, we still see a large drop from 69.23\% to 58.77\% (T\ref{table:refATHAblation},A4), and from 63.13\% to 53.78\% (T\ref{table:refATHAblation},A3). 

\noindent\textbf{Relationship detection (SGG3):}
Lastly, we integrate outputs from SGG3 into the scene graph to arrive at a fully generated scene graph. There is a similarly sized relative reduction in accuracy as seen for the introduction of SGG2: from 58.77\% to 51.71\% (T\ref{table:refATHAblation},A2) for Oracle op-seq, and 53.78\% to 47.55\% (T\ref{table:refATHAblation},ATH) to reach ATH's final overall accuracy without Oracle involvement.

\begin{table}
\begin{center}
\begin{tabular}{|l|c|c|c|c|c|c|}
\hline
System & QP & SG & Binary & Open & Grounding & Accuracy \\
\hline
ATH & ATH & ATH & 58.55 & 37.23 & 125.63 & 47.55 \\
\hline\hline
ATH-1 & ATH* & ATH & 63.31 & 37.96 & 126.79 & 50.22 \\
ATH-2 & GQA & ATH & 65.09 & 39.16 & 130.78 & 51.71 \\
ATH-3 & ATH & ATH obj+attr & 62.84 & 45.29 & 139.66 & 53.78 \\
ATH-4 & GQA & ATH obj+attr & 70.42 & 47.86 & 145.23 & 58.77 \\
ATH-5 & ATH & ATH obj & 72.63 & 54.22 & 139.38 & 63.13 \\
ATH-6 & GQA & ATH obj & 81.78 & 57.46 & 144.50 & 69.23 \\
ATH-7 & ATH & GQA & 78.55 & 82.44 & 165.62 & 80.56 \\
ATH-8 & ATH* & GQA & 84.65 & 84.76 & 167.91 & 84.71 \\
ATH-Oracle & GQA & GQA & 96.33 & 88.55 & 175.78 & 92.31 \\
\hline

\end{tabular}
\end{center}
\caption{ATH's performance for various combinations of using Oracle/predicted inputs. "GQA" entries stand for Oracle inputs from GQA annotations. "ATH*" skips ATH's learned QP but still goes through pre- and postprocessing which introduces some errors. "ATH-Oracle" represents ATH running with full Oracle input from GQA, which acts as the  upper bound of ATH. All results are based on GQA's balanced val set.}
\label{table:refATHAblation}
\end{table}

\subsection{Visual Grounding}
ATH produces clear inference paths on the scene graph for an input question. We use this path to compare ATH's visual grounding performance to four reference models: UpDn, MAC, N2NMN and PVR (see T\ref{table:rawAcc}).

\noindent\textbf{Metrics and Categories:} We use two metrics to measure grounding performance: 1) the official “Grounding” metric from GQA (code at  \cite{gqa_challenge}), and 2) an IoU-based metric to measure how much attention a model puts on objects referenced in GQA's grounding annotations (described below). We use two metrics here, because the GQA metric can create scores of >100\%, suggesting there might be an issue with the algorithm, but we want to be comparable to already reported numbers like \cite{pvr,gqa_dataset}, as shown in T\ref{table:rawAcc}.

We calculate the IoU-based grounding scores as follows:
First, we check for each GQA-annotated inference object if any scene graph object has an IoU of >0.5 with it. If yes, we add that scene graph object’s attention score to the final sum for that question.

GQA annotates inference objects for three categories: Objects referenced in the question (Q), in the short, one-word answer (A), and the full sentence answer (FA). We use these three and a 4th category with all three combined to calculate the IoU-based grounding scores as avg attention on inference objects per question.

We use a single attention map for all three compared models (UpDn, MAC, ATH). UpDn already produces a single attention map. MAC produces one map per step (4 in total), of which we take the average. For ATH, we assign an attention value to each traversed object in the final Viterbi path and normalize it by path length (e.g. if final path is [obj0 →  obj0 → obj3], attention distribution would be [obj0=0.66, obj3=0.33]).

\noindent\textbf{Results discussion:}
Grounding results for several systems reported with the "Grounding" metric are listed in T\ref{table:rawAcc}. ATH achieves by far the highest grounding score when compared to other systems. Evaluating IoU-based grounding scores for the categories (Q, A, FA) in more detail in T\ref{table:visualGrounding} shows that ATH features superior grounding compared to other models in all categories, improving scores between 12\% and 186\% relative compared to MAC. 
Results for various Oracle variants of ATH shows that its grounding improves further in particular with better scene graphs, while improvements in QP have less of an impact (e.g. going from ATH op-seq (T\ref{table:visualGrounding},A7) to Oracle op-seq (T\ref{table:visualGrounding},A-Oracle) show improvements of less than 10\% relative in all categories). This is consistent with trends we saw for accuracy (T\ref{table:refATHAblation}).

\begin{table}
\begin{center}
\begin{tabular}{|l||c|c||c|c|c|c||c|}
\hline
System & QP & SG & Q & A & FA & Q+A+FA & Grounding \\
\hline\hline
Det. Obj. & n/a & ATH & 91.45 & 92.34 & 91.66 & 91.49 & n/a \\
\hline\hline
UpDn & n/a & n/a & 15.60 & 22.05 & 17.71 & 19.39 & 69.95 \\
MAC & n/a & n/a & 18.48 & 25.35 & 21.16 & 22.05 & 86.75 \\
ATH & ATH & ATH & \textbf{52.87} & \textbf{28.29} & \textbf{50.58} & \textbf{56.16} & \textbf{125.63} \\
\hline\hline
ATH-2 & GQA & ATH & 56.25 & 29.41 & 53.32 & 59.34 & 130.78 \\
ATH-7 & ATH & GQA    & 73.17 & 49.33 & 73.35 & 81.25 & 165.62 \\
ATH-8 & ATH* & GQA    & 74.51 & 50.90 & 74.80 & 82.75 & 167.91 \\
ATH-Oracle & GQA & GQA    & 79.70 & 52.52 & 79.14 & 87.78 & 175.78 \\
\hline
Oracle & n/a  & n/a     & 100   & 100   & 100   & 100   & 197.03 \\
\hline
\end{tabular}
\end{center}
\caption{Visual grounding results. ATH shows superior grounding in all categories and both metrics. 
First line shows percentage of objects referenced in each category that were detected by ATH's SGG1. Last line shows numbers for Oracle attention input (i.e. perfect object selection and attention distribution).}
\label{table:visualGrounding}
\end{table}

\section{Conclusion}
We have introduced "Adventurer's Treasure Hunt" (ATH), a modular, transparent system based on strong visual grounding for the task of compositional VQA on scene graphs.

By its modular design, ATH exhibits great system transparency beneficial for error analysis and developing iterative improvements. In various evaluations we revealed that better object detection in particular shows the biggest potential for improving ATH's performance, while question parsing itself already performs very well in comparison.

In modeling the inference process after our "Adventurer's Treasure Hunt" analogy, we have shown that ATH achieves significantly better visual grounding in the inference process than reference systems like UpDn, MAC and PVR. 

Finally, with ATH, we have introduced the first GQA-based VQA system that does not use a restrictive, specially learned classifier for a pre-defined answer set in its answer generation process, but extracts answers dynamically based on the given scene graph, truly basing its answers on the contents of the knowledge base.

\bibliography{vqaib}
\end{document}